\documentclass[letterpaper, 10 pt, conference]{ieeeconf}  %

\IEEEoverridecommandlockouts        %
\usepackage[l2tabu,orthodox]{nag}

\usepackage[
    backend=bibtex8,
    style=ieee,
    sorting=none,
    natbib=true,
    doi=false,
    isbn=false,
    url=false,
    eprint=false,
    maxcitenames=1,
    mincitenames=1
]{biblatex}

\usepackage[pdftex,colorlinks]{hyperref}

\usepackage[printonlyused]{acronym}

\usepackage{siunitx}
\usepackage[all]{nowidow}

\usepackage[dvipsnames]{xcolor}

\usepackage{lipsum}

\usepackage{xspace} %

\usepackage[pdftex]{graphicx}

\usepackage{epstopdf}

\usepackage{import}

\graphicspath{{./latexGoodPractices/}}

\usepackage{booktabs}

\usepackage{amssymb,amsfonts,amsmath,amscd}

\usepackage{bm}

\newcommand{\abs}[1]{\left\vert#1\right\vert}

\newcommand{\bbm}{\begin{bmatrix}}
        \newcommand{\ebm}{\end{bmatrix}}

\usepackage[utf8]{inputenc}

\usepackage{siunitx}
\usepackage{setspace}

\usepackage{datatool}
\DTLsetseparator{ = }
\newcommand{\data}[2]{\DTLgetvalueforkey\currentvalue{thevalue}{#1}{thekey}{#2}\currentvalue}
\newcommand{\numdata}[2]{\DTLgetvalueforkey{\currentvalue}{thevalue}{#1}{thekey}{#2}\num{\currentvalue}}

\newcommand{\bnumdata}[2]{%
    \renewrobustcmd{\bfseries}{\fontseries{b}\selectfont}%
    \sisetup{detect-weight,mode=text}%
    \DTLgetvalueforkey{\currentvalue}{thevalue}{#1}{thekey}{#2}\textbf{\num{\currentvalue}}%
}

\usepackage{amsmath,amsfonts,amssymb}
\usepackage{mathtools}
\usepackage{cancel}
\usepackage[scr]{rsfso}
\usepackage{nicefrac, xfrac}

\usepackage{pgfplots}
\pgfplotsset{compat=1.15}

\usepackage{graphicx}
\usepackage{float}
\usepackage{subcaption}
\usepackage[rightcaption]{sidecap}

\usepackage[printonlyused]{acronym}

\usepackage{booktabs}
\usepackage{multirow,multicol,makecell}
\usepackage{tabu}

\usepackage{array}
\newcolumntype{Y}{>{\centering\arraybackslash}X}
\newcolumntype{M}[1]{>{\centering\arraybackslash}m{#1}}
\newcolumntype{P}[1]{>{\centering\arraybackslash}p{#1}}
\newcolumntype{L}[1]{>{\raggedleft\arraybackslash} p{#1}}
\newcolumntype{R}[1]{>{\raggedright\arraybackslash}p{#1}}

\newcolumntype{C}[1]{>{\centering\arraybackslash}m{#1}}

\usepackage{circuitikz,tikz}
\usetikzlibrary{positioning,calc,arrows.meta,math,shapes.misc,plotmarks}

\usepackage{listings}

\usepackage{fancyvrb}
\usepackage{fvextra}

\RecustomVerbatimCommand{\VerbatimInput}{VerbatimInput}%
{%
    fontsize=\footnotesize,%
    frame=lines,  %
    rulecolor=LightGray
}

\usepackage{csquotes}

\usepackage{url}
\usepackage{bookmark}

\usepackage[french,english,noabbrev,nameinlink,capitalise]{cleveref}

\usepackage[intoc]{nomencl}
\makenomenclature

\usepackage{etoolbox}
\ExplSyntaxOn
\NewExpandableDocumentCommand{\strcase}{mm}
{
    \str_case:nn { #1 } { #2 }
}
\ExplSyntaxOff

\usepackage{ifthen}

\usepackage[inkscapearea=page]{svg}
\usepackage{balance}    %
\usepackage{soul}  %
\usepackage{textpos}    %
\usepackage{rotating}   %
\usepackage[normalem]{ulem}
\usepackage{xcolor}

\crefname{subfigure}{Figure}{Figures}
\crefname{figure}{Figure}{Figures}
\crefname{table}{Table}{Tables}

\acrodef{UGV}{uncrewed ground vehicle}
\acrodefplural{UGV}{uncrewed ground vehicles}
\acrodef{IMU}{Inertial Measurement Unit}
\acrodef{SSMR}{skid-steering mobile robot}
\acrodefplural{SSMR}{skid-steering mobile robots}
\acrodef{TnR}[T\&R]{Teach and Repeat}
\acrodef{ROS}{Robot Operating System}
\acrodef{SOTA}{state-of-the-art}
\acrodef{SLAM}{Simultaneous Localization and Mapping}
\acrodef{DOF}{Degrees Of Freedom}
\acrodef{EKF}{Extended Kalman Filter}
\acrodef{ICR}{instantaneous center of rotation}
\acrodefplural{ICR}{instantaneous centers of rotation}
\acrodef{MMP}{\textit{main motion primitive}}
\acrodefplural{MMP}{\textit{main motion primitives}}
\acrodef{RLS}{recursive least squares}
\acrodef{IQR}{interquartile range}
\acrodef{TC}{terrain classification}

\acrodef{FFT}{fast Fourier transform}
\acrodefplural{FFT}{fast Fourier transforms}
\acrodefplural{DFT}{discrete Fourier transform}
\acrodef{PSD}{power spectral density}
\acrodefplural{PSD}{power spectral densities}
\acrodef{MFCC}{Mel-frequency cepstral coefficient}
\acrodefplural{MFCC}{Mel-frequency cepstral coefficients}
\acrodef{STFT}{short-time Fourier transform}

\acrodef{ML}{Machine Learning}
\acrodef{MLP}{Multilayer Perceptron}
\acrodef{RF}{Random Forest}
\acrodef{SVM}{Support Vector Machine}
\acrodef{SMOTE}{Synthetic Minority Over-sampling Technique}
\acrodef{OOB}{out-of-bag}
\acrodef{DT}{Decision Tree}
\acrodefplural{MLP}{Multilayer Perceptrons}
\acrodefplural{RF}{Random Forests}
\acrodefplural{DT}{Decision Trees}
\acrodef{ANN}{Artificial Neural Network}
\acrodef{RNN}{Recurrent Neural Network}
\acrodef{ReLU}{Rectified Linear Unit}
\acrodef{LSTM}{Long Short-Term Memory}
\acrodef{CNN}{Convolutional Neural Network}
\acrodef{TL}{Transfer Learning}
\acrodef{RBF}{radial basis function}
\acrodef{SSM}{state space model}
\acrodefplural{SSM}{state space models}
\acrodef{t-SNE}{t-distributed stochastic neighbor embedding}
\acrodef{S4}{Structured State Space for Sequence Modeling}
\acrodef{C-LSTM}{Convolutional LSTM}
\acrodef{VAE}{Variational Auto-Encoder}
\acrodef{GMM}{Gaussian Mixture Model}
\acrodef{BN}{batch normalization}
\acrodef{NLP}{natural language processing}
\acrodef{k-NN}[\textit{k}-NN]{\textit{k}-nearest neighbors}

\acrodef{SC}{San Cassiano}
\def\HuskyData{\texttt{BorealTC}}
\def\VulpiData{\texttt{Vulpi}}

\acrodef{DC}{direct current}
\acrodef{BDC}{brushed DC}
\acrodef{BLDC}{brushless DC}

\newcommand{\motvar}[3]{%
    \ifthenelse{\equal{#2}{}}
    {\ensuremath{{#1}^{\text{#3}}}}
    {\ifthenelse{\equal{#3}{}}
        {\ensuremath{{#1}_{\text{#2}}}}
        {\ensuremath{{#1}_{\text{#2}}^{\text{#3}}}}}
}

\DeclareSIUnit{\pp}{\%pt}

\title{\LARGE \bfseries
    Proprioception Is All You Need: Terrain Classification for Boreal Forests
}

\author{Damien LaRocque,$^{1}$ William Guimont-Martin,$^{1}$ David-Alexandre Duclos,$^{1}$\\Philippe Giguère,$^{1}$ François Pomerleau$^{1}$
\thanks{
    \raggedright$^{1}$ The authors are with the Northern Robotics Laboratory, Université Laval, Quebec City, Canada,
    {\texttt{\small{damien.larocque@norlab.ulaval.ca, francois.pomerleau@norlab.ulaval.ca}}}
}%
}

\usepackage{tikz}
\newcommand\copyrighttext{%
	\footnotesize \textcopyright 2024 IEEE. Personal use of this material is permitted. Permission from IEEE must be obtained for all other uses, in any current or future media, including reprinting/republishing this material for advertising or promotional purposes, creating new collective works, for resale or redistribution to servers or lists, or reuse of any copyrighted component of this work in other works.}
\newcommand\copyrightnotice{%
	\begin{tikzpicture}[remember picture,overlay]
		\node[anchor=south,yshift=15pt] at (current page.south) {\parbox{\dimexpr\textwidth-\fboxsep-\fboxrule\relax}{\copyrighttext}};
	\end{tikzpicture}%
}

\begin{document}

\maketitle
\copyrightnotice
\thispagestyle{empty}
\pagestyle{empty}

\begin{abstract}

  Recent works in field robotics highlighted the importance of resiliency against different types of terrains.
  Boreal forests, in particular, are home to many mobility-impeding terrains that should be considered for off-road autonomous navigation.
  Also, being one of the largest land biomes on Earth, boreal forests are an area where autonomous vehicles are expected to become increasingly common.
  In this paper, we address the issue of classifying boreal terrains by introducing \HuskyData{}, a publicly available dataset for proprioceptive-based \ac{TC}.
  Recorded with a \textit{Husky} A200, our dataset contains \SI{116}{\minute} of \ac{IMU}, motor current, and wheel odometry data, focusing on typical boreal forest terrains, notably snow, ice, and silty loam.
  Combining our dataset with another dataset from the literature, we evaluate both a \ac{CNN} and the novel \ac{SSM}-based Mamba architecture on a \ac{TC} task.
  We show that while \ac{CNN} outperforms Mamba on each separate dataset, Mamba achieves greater accuracy when trained on a combination of both.
  In addition, we demonstrate that Mamba's learning capacity is greater than a \ac{CNN} for increasing amounts of data.
  We show that the combination of two \ac{TC} datasets yields a latent space that can be interpreted with the properties of the terrains.
  We also discuss the implications of merging datasets on classification.
  Our source code and dataset are publicly available online: \url{https://github.com/norlab-ulaval/BorealTC}.
\end{abstract}

\acresetall
\section{Introduction}\label{sec:intro}

With the ongoing development of field robotics, it has become common for robots to navigate through increasingly complex and challenging terrains~\cite{Borges2022}.
To prevent and handle situations where an \acf{UGV} may get stuck or immobilized, vehicles must be able to accurately assess and identify the terrain they are navigating on.
Such terrain awareness is often framed as a classification problem over the different terrain types a \ac{UGV} might traverse~\cite{Zurn2021,Vulpi2021}.
The problem of \acf{TC} has been applied in many contexts, including traversability assessment~\cite{Borges2022}, terrain-aware path planning~\cite{Atha2022}, and as a prior for predicting energy consumption~\cite{Allred2022}. %

Although being the largest land biome on Earth~\cite{Hayes2022Boreal}, boreal forests have received little attention for the development of autonomous navigation.
Moreover, terrain awareness is essential in the context of the boreal forest, where a multitude of terrain types can significantly hinder the mobility of \acp{UGV}~\cite{Baril2022}.
Subject to large seasonal variability, boreal forests are especially suitable for developing systems that are capable of multi-seasonal navigation~\cite{Ali2020}.
Hence, \ac{TC} in these regions is essential to the advancement of field robotics in more challenging conditions.

\begin{figure}[t]
  \centering
  \includegraphics[width=\linewidth]{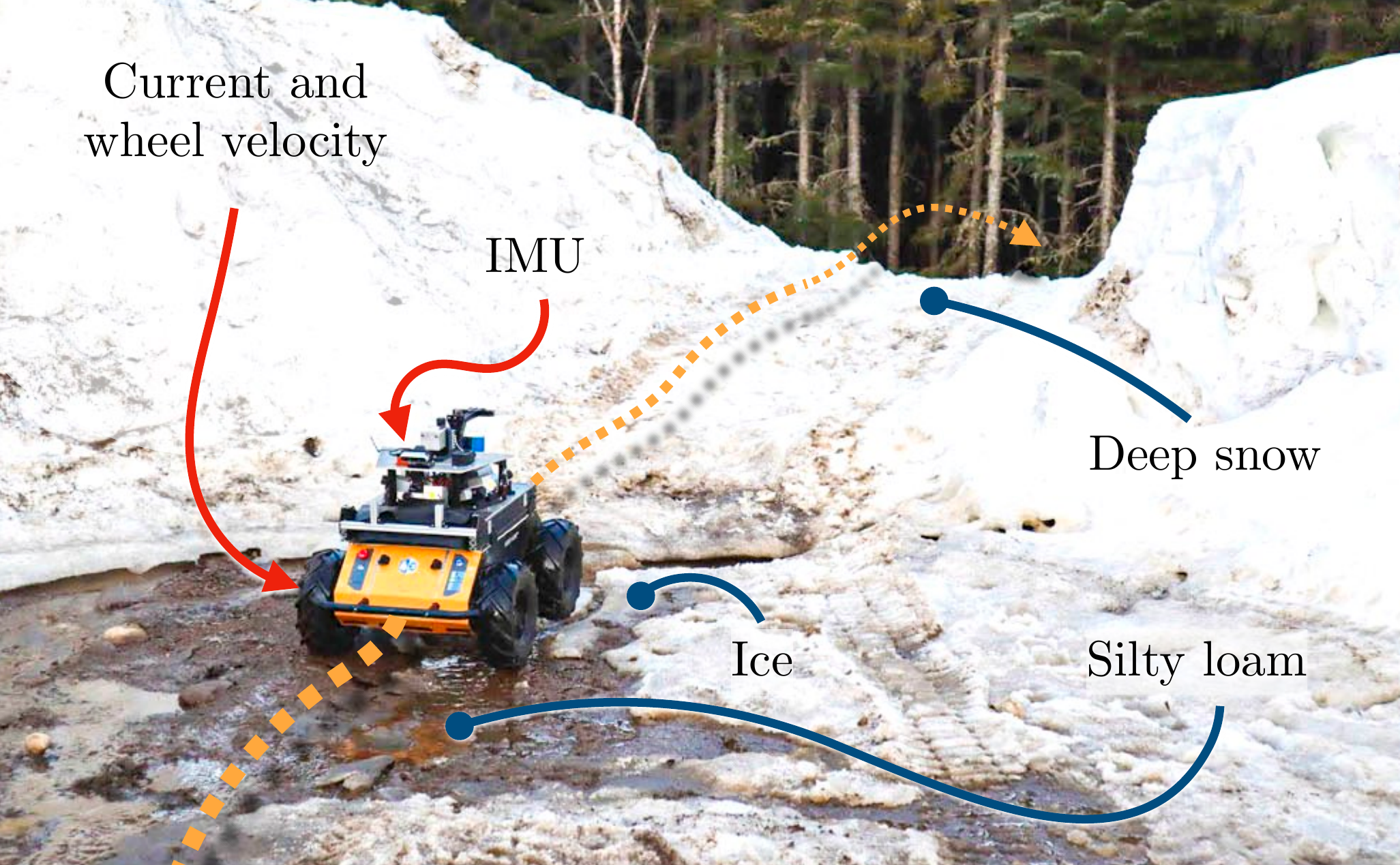}
  \caption{An example of challenges caused by terrain in boreal forests.
    A small \ac{UGV} with a controller relying on a friction coefficient expecting concrete will struggle to follow a path when confronted with complex ground-wheel interactions.
  }
  \label{fig:husky-in-context}
  \vspace{-0.75\baselineskip}
\end{figure}

Although a traversed terrain can be determined with cameras~\cite{Zurn2021}, this approach is not universal for in-situ \ac{TC}, depending on environmental conditions.
For example, boreal forests challenge conventional visual-based \ac{TC}, as their dense coniferous canopies obstruct the sunlight, yielding low visual feature contrast~\cite{Baril2022}.
Furthermore, boreal forests are in regions with large illumination variances, having shorter days during winter.
Hence, a \ac{TC} method that relies on light does not work anytime in winter when light is usually scarce~\cite{Paton2016}.
While lidars can be used in low lighting conditions, they provide little semantic information about the nature of the surface, given they only provide geometric information from the surroundings~\cite{Schilling2017}.
Both cameras and lidars are influenced by adverse environmental conditions such as snowstorms~\cite{Courcelle2023}, heavy rain, or fog~\cite{Williams2020}.
Finally, as illustrated in \cref{fig:husky-in-context}, the terrain configuration in boreal forests can be quite complex, and only a thin layer of snow accumulation can hinder the capacity of visual sensors to see the ground composition.
This occlusion is particularly true when a fresh layer of snow covers loam or ice, derailing controllers that assume a constant friction coefficient~\cite{Baril2024}.
In this context, proprioceptive sensing is more robust for \ac{TC} in harsh conditions~\cite{Vulpi2021}, such as those found in boreal forests.

For practical reasons, many authors have gathered terrain data indoors, in urban settings, or on university campuses, where terrain usually does not impede movement~\cite{Ishikawa2021,Zurn2021,Allred2022}.
In contrast, our work focuses primarily on off-road terrains that pose challenges for wheeled \acp{SSMR}.
In particular, we consider deep snow, known for immobilizing wheeled \acp{SSMR}~\cite{Baril2022}; ice, a rigid slippery surface; and silty loam, a muddy and slippery soil that hampers rotations.

Current approaches in \ac{TC} generally incorporate temporal aspects, either through frequency domain representations or by employing recurrent models.
Building upon these foundations, we introduce two models.
Our first model improves upon previous \acf{CNN} baseline architectures applied to spectrograms generated from proprioceptive sensor data.
We employ recent deep learning techniques and integrate windowing functions when computing spectrograms to improve the model's performance.
The second model draws inspiration from recent advancements in \acfp{SSM}, specifically the Mamba architecture introduced by \citet{Gu2023Mamba}.
Mamba is a promising approach for \ac{TC}, due to its ability to apply selectivity on raw sequential data while scaling linearly with sequence length.

The main contributions of this paper are
(i)~\HuskyData{}, a novel dataset for \ac{TC} with wheeled \acp{SSMR} in wintry off-road conditions;
(ii) an improvement on state-of-the-art methods for data-driven \ac{TC} based on \acp{CNN};
(iii) the exploration of \ac{SSM}-based models for \ac{TC}; and
(iv) a study of the challenges of merging %
datasets acquired by different \acp{UGV} of the same model.

\section{Related Work}\label{sec:RW}

\subsection{Sensor modalities}

Exteroceptive sensors have been extensively used for \acf{TC}, as they can predict terrains at a distance.
Among these sensors, cameras stand out for their ability to capture appearance-based features, providing valuable insights into the physical nature of surfaces.
For example, \citet{Atha2022} classified terrain from Martian rovers' mast camera images.
\citet{Walas2014} classified terrains from lidar data, by creating intensity-based and geometry-based feature vectors.
\ac{TC} can also benefit from the combination of sensors employing different modalities.
For instance, \citet{Schilling2017} leveraged both geometric and appearance-based features from lidar and camera data to assess terrain traversability.
Audio signals have also been combined with camera images~\cite{Ishikawa2021,Zurn2021} and radar scans~\cite{Williams2020} for \ac{TC}.
Proprioceptive data were used with camera images to classify terrain in both agricultural~\cite{Reina2017} and urban contexts~\cite{Chen2021}.
While exteroceptive sensors offer many advantages, they can be significantly challenged in boreal forests.
For instance, cameras will suffer from illumination variability~\cite{Paton2016,Baril2022}.
Moreover, lidars are affected by inclement weather and extreme precipitation~\cite{Williams2020,Courcelle2023}, typical of the same region.

To circumvent these limitations, an alternative approach is to base \acl{TC} primarily on proprioceptive data.
Proprioceptive sensors present the advantage of directly informing about the physical characteristics of a surface through their impact on the dynamics of a \ac{UGV}.
Hence, they do not require an unobstructed line of sight with the surface, nor do they rely on surface illumination.
Common proprioceptive sensors are \acfp{IMU}, wheel odometry, and motor ammeters, with the latter providing indirect torque measurements.
\acp{IMU} yield accelerations that can be used to classify terrains.
These measurements are especially useful for legged robots, where the body is far from the ground~\cite{Giguere2009,Otsu2016}.
For finer surface information, the accelerometer can be dragged on the ground~\cite{Giguere2011}, bypassing the damping effect of legs or wheels.
In addition, actuator and haptic signals can be leveraged for \ac{TC} when dealing with legged robots.
For example, leg force measurements from the \textit{Messor} walking machine~\cite{Dallaire2015} or leg haptic signals from an \textit{ANYmal} walking robot~\cite{Bednarek2022} can provide valuable information for classification.
\citet{Allred2022} also classified terrain using leg joints data, in conjunction with images of the front-facing camera of a Spot robot.
For wheeled robots with encoders and \ac{IMU}, \citet{Reina2016} have shown that proprioceptive data can be used to evaluate slip and motion resistance coefficients to predict the terrain on which a \ac{UGV} was driven.
By adding motor currents to wheel velocities, and \ac{IMU}, \citet{Vulpi2021} demonstrated a mean accuracy of \SI{91.5}{\%} for \ac{TC} over four types of terrain.
Given that relying on proprioceptive sensors to record terrain signatures will be strongly related to the sensor placement, the geometry of the robot, and the type of locomotion, we will use their data in combination with ours to analyze \emph{domain shift}, which makes knowledge transfer between vehicles challenging.

\subsection{Methods for classification}

Earliest approaches involved expert systems~\cite{Reina2016} and \acp{SVM} for lidar-based~\cite{Walas2014} and vibration-based \acf{TC}~\cite{Otsu2016,Reina2017}.
Yet, this family of \ac{ML} techniques has the drawback of relying on features that were hand-crafted with a priori expert knowledge, which adds an inductive bias to the learning.

More recently, deep learning approaches have gained popularity due to their representation learning capabilities and their ability to process any type of sensor information.
For instance, \ac{CNN} architectures have been used to classify terrains based on camera images~\cite{Atha2022,Allred2022,Chen2021}.
In the case of~\cite{Chen2021}, the incorporation of proprioceptive data through a second parallel network improved classification accuracy.

Inspired by speech recognition applications, 1-D data, such as \ac{IMU}-recorded vibrations and audio signals, are often transformed into frequency representations.
One example of this technique is the utilization of \ac{STFT}-based spectrograms by \citet{Vulpi2021}, who employed a \ac{CNN} to classify terrains with proprioceptive data from an \ac{IMU} and the drive system of a \ac{UGV}.
Similarly, \citet{Zurn2021} applied the same type of spectrogram for unsupervised acoustic feature learning.
These learned acoustic features are part of a self-supervised framework for audiovisual-based \ac{TC} using neural encoders.
Likewise, \citet{Ishikawa2021} employed \acp{VAE} and a \ac{GMM} to learn terrain types from audiovisual data autonomously.
In their approach, the audio signals were represented as \acp{MFCC}, again inspired by speech recognition techniques.
Building upon these methods, our \ac{CNN} classifier uses a \ac{STFT}-based spectrogram.
In contrast to \citet{Vulpi2021}, we demonstrate that applying a windowing function mitigates spectral leakage.

Another way to process 1-D time series is to input them into a neural network designed for sequential data.
In such cases, recurrent networks like \acp{LSTM} are commonly used.
\citet{Allred2022} achieved high accuracy on \ac{TC} by applying a \ac{LSTM} on the joint data of a legged robot.
A more complex variant of \acp{RNN}, a convolutional \ac{LSTM} (C-LSTM), was used by \citet{Valada2017IJRR} for audio-based \ac{TC} on \acp{MFCC}.
As \citet{Vulpi2021} demonstrated, the \ac{CNN} architecture is more stable and more accurate than \ac{LSTM} and C-LSTM for proprioceptive-based \ac{TC} on a wheeled \ac{UGV}, hence these methods won't be investigated.

In addition to \ac{LSTM}-based approaches, other methods inspired by \ac{NLP} have emerged for \ac{TC}.
Transformers, in particular, have recently been proposed to process long sequences and classify terrains as accurately as \acp{RNN}~\cite{Bednarek2022}.
However, the performance of transformer-based approaches comes at the expense of quadratic scaling, in proportion to the sequence length.
To address this limitation, other \ac{RNN}-like approaches have been suggested.
Most notably, \citet{Gu2023Mamba} introduced Mamba, a \ac{SSM}-based architecture that offers significant performance gains while scaling linearly.
While Mamba aligns with the recurrent nature of \acp{RNN}, it distinguishes itself by employing selectivity, emphasizing key time steps in data sequences while leveraging its operations' parallelism.
Although Mamba has been shown to beat recent architectures in various downstream tasks such as DNA sequence classification~\cite{Gu2023Mamba}, its application to \ac{TC} remains unexplored.
Therefore, we evaluate the potential of Mamba for this task.

\subsection{Datasets for \acl{TC}}

For practical reasons, some studies have focused their efforts on classifying data acquired indoor~\cite{Walas2014,Dallaire2015,Bednarek2022} or in urban environments~\cite{Allred2022,Valada2017IJRR,Ishikawa2021}.
For example, wheeled \acp{UGV} were recently used for multi-modal data acquisition in urban areas, such as for the \textit{Freiburg Terrains} and the \textit{Jackal robot 7-class terrain} datasets~\cite{Zurn2021,Chen2021}.
\Acf{TC} was explored in various off-road contexts, such as an experimental farm~\cite{Reina2016,Reina2017,Vulpi2021}, a volcanic island~\cite{Otsu2016}, and Mars~\cite{Atha2022}.
Given that the boreal forest remains covered in snow for at least half of the year, any dataset covering such an environment ought to include data on snow and ice.
These two wintry terrains have been studied with legged robots~\cite{Giguere2009,Allred2022}.
However, to the best of our knowledge, no publicly available \ac{TC} datasets contain labeled snow and ice data from a wheeled \ac{UGV}.

Many have publicly released their datasets to help generalize classification across different kinds of terrains~\cite{Chen2021,Vulpi2021,Zurn2021,Allred2022,Atha2022,Bednarek2022}.
Aligned with this endeavor, we propose \HuskyData{}, a publicly available dataset containing annotated data from a wheeled \ac{UGV} for various mobility-impeding terrain types typical of the boreal forest.
Our data were acquired on deep snow and silty loam, two uncommon terrains in an urban setting, both in winter and spring.
Additionally, to encompass a variety of wintry terrains, we recorded data while driving our \ac{UGV} on an ice rink.
Our dataset also includes experiments on asphalt and flooring, two prevalent terrains in recent datasets.
These types of terrain facilitate the comparison of the learned representations of our models with those obtained from other datasets.

\section{Methodology}\label{sec:method}

\begin{figure*}[htbp]
  \centering
  \includegraphics[width=0.99\linewidth]{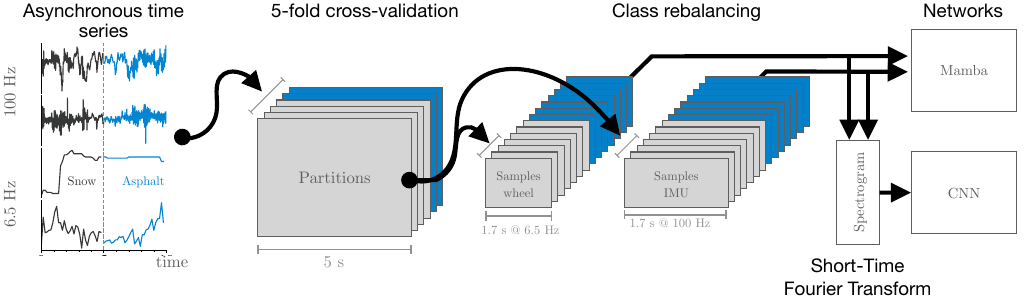}
  \caption{Overview of the training process.
  From the left, data from asynchronous sensors were recorded and hand-labeled according to the terrain on which the robot was driven.
  To allow a 5-fold cross-validation, trajectories are split into \SI{5}{\second} partitions.
  Classes are then rebalanced through oversampling before being fed to the different networks.
  The \ac{CNN} performed classification on spectrograms, while Mamba classified the samples directly in the time domain.
  }~\label{fig:flowchart}
  \vspace{-1.75\baselineskip}
\end{figure*}

We propose an approach to classify terrains using proprioceptive data from an \ac{IMU}, and the drive system of a \ac{UGV}.
\cref{fig:flowchart} shows the general overview of the pipeline used for the evaluation.
Following \citet{Vulpi2021}, we divided our sensor signals into \SI{5}{\second} partitions.
These partitions were then split into train and test subsets, to evaluate the performance of our models with a \textit{k}-fold cross-validation strategy.
To overcome the class imbalance in the data, the partitions in both subsets were oversampled, such that all classes have the same number of samples.
Subsequently, two models were applied to each sample: a \ac{CNN} and a Mamba classifier. %

\subsection{Convolutional Neural Network}\label{sec:CNN}

Following the method of \citet{Vulpi2021}, we apply a \ac{CNN} to spectrograms generated from proprioceptive data.
The spectrograms were computed on each \SI{1.7}{\second} sample by applying \acp{STFT} with a window length of \SI{0.4}{\second} and an overlap of \SI{0.2}{\second}.
The original implementation~\cite{Vulpi2021} applied the \ac{STFT} directly on these small windows, which is equivalent to applying boxcar filters.
This filter results in spectra with artifacts, called \emph{spectral leakage}, that must be avoided~\cite{Harris1978}.
Instead, we use a Hamming window \(w\left[n\right]\), such that
\begin{equation}
  w\left[n\right] = a_0 - (1-a_0) \cdot \cos\left(\frac{2\pi{n}}{N_{win}}\right), \label{eq:hamming-window}
\end{equation}
for \(0 \leq n \leq N_{win}\), with $N_{win}$ being the number of samples per window in the \ac{STFT} and \(a_0=0.54\)~\cite{Blackman1958}.
As in \citet{Vulpi2021}, the resulting spectrograms of each channel were then padded, concatenated, and fed as an input to the \ac{CNN}.
The convolution operations of the \ac{CNN} were performed across all channels of the spectrograms, by moving the kernel through the frequency-time planes.

\subsection{Mamba}\label{sec:mamba}

In light of state-of-the-art results on multiple tasks involving sequential data, we suggest using the Mamba architecture~\cite{Gu2023Mamba} for \acf{TC}.
Based on recent work with discrete \acp{SSM}, Mamba introduces attention-like selectivity and recurrent-like parallel associative scanning for linear scaling in sequence length.
Notably, Mamba obviates the necessity for domain-shifting the samples to spectrograms, and consequently the need for preprocessing steps, like padding or downsampling, required by \acp{CNN} for data uniformity.
Mamba thus directly processes the proprioceptive data in its sequential form, making it a promising solution for proprioceptive-based \ac{TC}.

\section{Experiments}\label{sec:experiments}
In this section, we first describe the platform used to record the \HuskyData{} dataset.
We then give details about the terrain and highlight differences between ours and \VulpiData{} dataset.
We finally give implementation details for both \ac{ML} architectures used in our analyses.

\subsection{Vehicle and sensors}\label{sec:vehicle}

Our dataset was recorded with a \textit{Husky} A200, a wheeled \ac{UGV} from \textit{Clearpath Robotics} (Kitchener, Ontario, Canada).
The \textit{Husky} is a \ac{SSMR} with a weight of \SI{70}{\kilogram} and a wheel baseline of \SI{0.6}{\metre}.
While the \ac{UGV} has four wheels, the wheels are mechanically coupled with timing belts, such that a single gearmotor drives two wheels on the same side of the vehicle.
The motor currents are measured by MDL-BDC24 motor drivers, while the wheel speeds are collected using optical encoders mounted at the output of the gearmotors.
Both currents and wheel velocities are provided by the \textit{Husky} at a rate of \SI{6.5}{\hertz}.
The robot is also equipped with a Xsens MTi-30 \ac{IMU}, which records three angular velocities and three linear accelerations at a frequency of \SI{100}{\hertz}.
To ensure consistency with the data of \citet{Vulpi2021}, the \ac{IMU} data was transformed to the base reference frame of the robot, using the Coriolis formula as described by \citet{Deschenes2024}.
Finally, all sensor data was recorded with \ac{ROS} 2.
The robot and its sensors are shown in \cref{fig:husky-in-context}.

\subsection{Dataset description}\label{sec:dataset}

\begin{figure*}[htbp]
  \centering
  \captionsetup[subfigure]{belowskip=-1pt}
  \begin{subfigure}[b]{0.19\linewidth}
    \includegraphics[width=\linewidth]{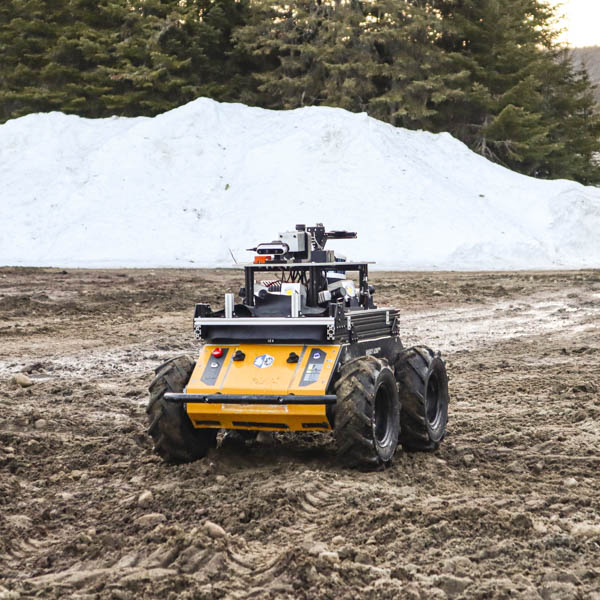}
  \end{subfigure} \hfill
  \begin{subfigure}[b]{0.19\linewidth}
    \includegraphics[width=\linewidth]{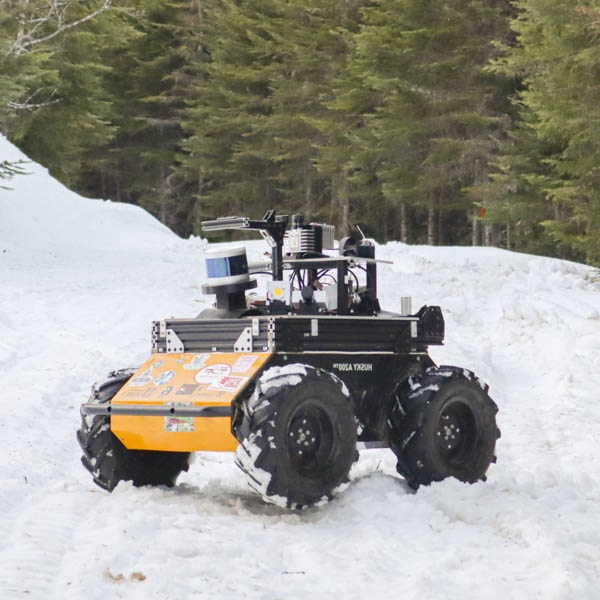}
  \end{subfigure} \hfill
  \begin{subfigure}[b]{0.19\linewidth}
    \includegraphics[width=\linewidth]{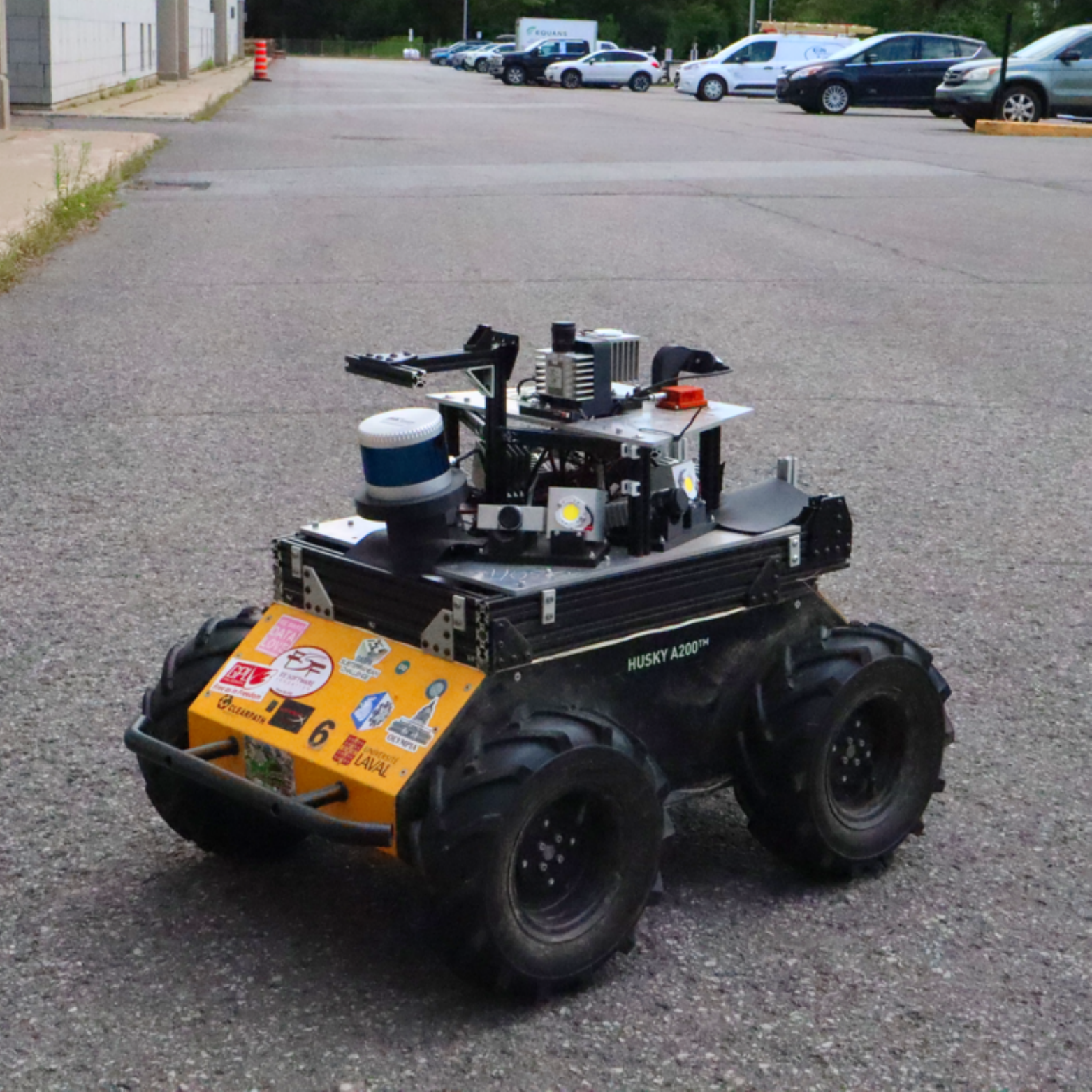}
  \end{subfigure} \hfill
  \begin{subfigure}[b]{0.19\linewidth}
    \includegraphics[width=\linewidth]{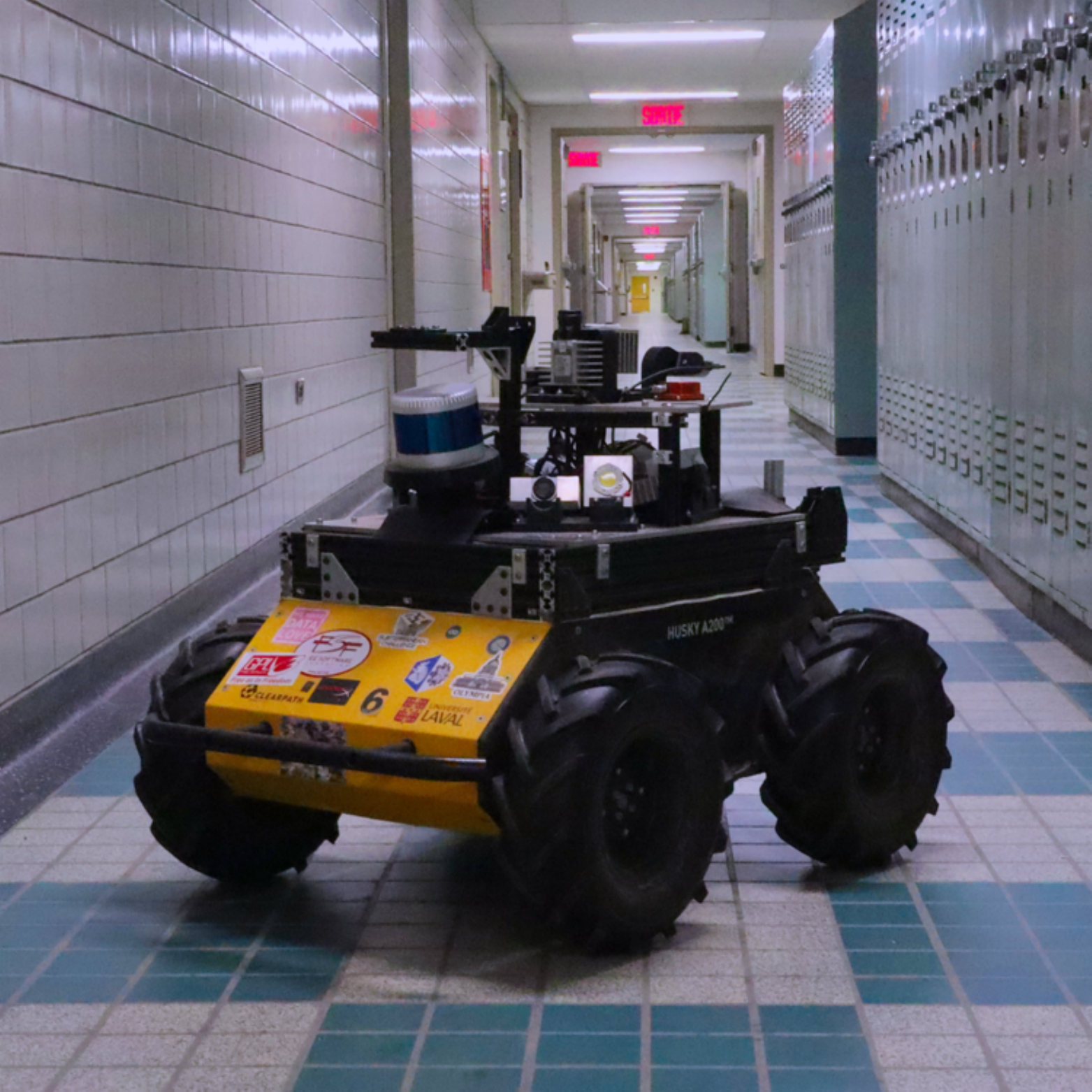}
  \end{subfigure} \hfill
  \begin{subfigure}[b]{0.19\linewidth}
    \includegraphics[width=\linewidth]{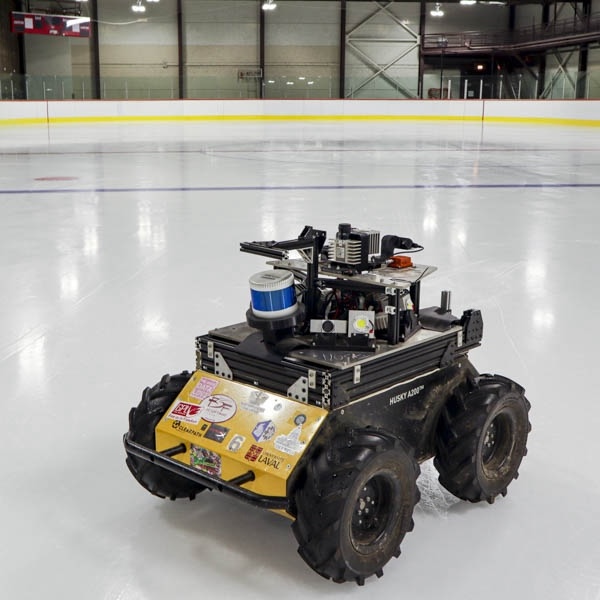}
  \end{subfigure}
  \caption{Types of terrains considered in our dataset.
  From left to right: silty loam, deep snow, asphalt, flooring, and ice.
  }~\label{fig:terrains}
  \vspace{-1.75\baselineskip}
\end{figure*}

Our \HuskyData{} dataset was collected by driving the \textit{Husky} on five different types of terrains, namely \textsc{asphalt}, \textsc{flooring}, \textsc{ice}, \textsc{silty loam}, and \textsc{snow}, as shown in \cref{fig:terrains}.
\textsc{Asphalt}, \textsc{flooring}, and \textsc{ice} data were acquired in an urban setting, on the campus of \emph{Université Laval} (\ang{46;46;52.47}N,~\ang{71;16;27.74}W).
\textsc{Ice} data was recorded on an ice rink on the same campus, ensuring that the ice was consistent between experiments.
\textsc{Silty loam} and \textsc{snow} data were taken at \textit{Forêt Montmorency} (\ang{47;19;19.29}N,~\ang{71;8;50.13}W), the experimental boreal forest of \emph{Université Laval}, \SI{75}{\kilo\metre} north of its main campus.
During all seasons, excluding winter, the soil of \textit{Forêt Montmorency} is a podzol typical of boreal forests.
More specifically, the trails on which the \textit{Husky} was driven are dug in the silty loam layer of the podzol.
As we recorded the \textsc{silty loam} data at the end of the winter, the silty loam was saturated in water, making it slippery.
All the data were collected on relatively flat surfaces, with a pitch smaller than \SI{5}{\degree}, to avoid effects from the slope of the terrain on the classification.

To better evaluate our methods, we used our \HuskyData{} dataset in conjunction with the \VulpiData{} dataset~\cite{Vulpi2021}.
As the data follows the work of \citet{Reina2017}, which was recorded in San Cassiano, Lecce, Italy (\ang{40;03;35.40}N,~\ang{18;20;50.98}E), we extrapolate that the dataset comes from the same location.
As with our dataset, the \VulpiData{} dataset was collected with a \textit{Clearpath Robotics} \textit{Husky}, meaning that both datasets were acquired with vehicles of similar dimensions and weights.
While the acquisition rates of our \ac{IMU} and wheel data are at \SI{100}{\hertz} and \SI{6.5}{\hertz} respectively, the \VulpiData{} dataset has a rate of \SI{50}{\hertz} for their \ac{IMU} and \SI{15}{\hertz} for their wheel data.
A summary of both \HuskyData{} and \VulpiData{} datasets is given in \cref{tab:dataset}, where the ranges of motion commands that were sent to the \acp{UGV} are compared for each class.
This comparison is done by computing the median \(\tilde{\cdot}\) and the \ac{IQR} of the absolute values \(\abs{v_x}\) and \(\abs{\omega_z}\)  of the linear and angular velocities.
Notably, our \HuskyData{} dataset is an order of magnitude larger and contains a total of \SI{116}{\minute} of sensor data, while \VulpiData{} contains \SI{13}{\minute}.
In addition, our dataset includes a significant amount of data with rotational motions, such as turns-on-the-spot, while the \VulpiData{} dataset contains mostly forward linear motions.
The inclusion of rotational motions enables a better representation of the entire input space of a \ac{UGV}~\cite{Baril2024}, and is thus crucial for accurate terrain modeling.

\begin{filecontents*}{data-vulpi-terrains.dat}
CONCRETE = 24
DIRT_ROAD = 16
PLOUGHED = 60
UNPLOUGHED = 56
\end{filecontents*}
\DTLloaddb[noheader, keys={thekey,thevalue}]{vterr}{data-vulpi-terrains.dat}
\begin{filecontents*}{data-norlab-terrains.dat}
ASPHALT = 111
FLOORING = 423
ICE = 450
SANDY_LOAM = 126
SNOW = 281
\end{filecontents*}
\DTLloaddb[noheader, keys={thekey,thevalue}]{nterr}{data-norlab-terrains.dat}
\begin{filecontents*}{data-vulpi-stats.dat}
iw-DIR = 0.00
mw-DIR = 0.00
iv-DIR = 0.25
mv-DIR = 0.56
iw-PLO = 0.00
mw-PLO = 0.00
iv-PLO = 0.26
mv-PLO = 0.56
iw-CON = 0.00
mw-CON = 0.00
iv-CON = 0.26
mv-CON = 0.56
iw-UNP = 0.00
mw-UNP = 0.00
iv-UNP = 0.25
mv-UNP = 0.56
\end{filecontents*}
\DTLloaddb[noheader, keys={thekey,thevalue}]{vstat}{data-vulpi-stats.dat}
\begin{filecontents*}{data-husky-stats.dat}
iw-ASP = 0.09
mw-ASP = 0.01
iv-ASP = 0.58
mv-ASP = 0.46
iw-SAN = 0.17
mw-SAN = 0.10
iv-SAN = 0.24
mv-SAN = 0.00
iw-ICE = 0.52
mw-ICE = 0.27
iv-ICE = 0.38
mv-ICE = 0.24
iw-FLO = 0.09
mw-FLO = 0.02
iv-FLO = 0.05
mv-FLO = 0.23
iw-SNO = 0.26
mw-SNO = 0.10
iv-SNO = 0.31
mv-SNO = 0.00
\end{filecontents*}
\DTLloaddb[noheader, keys={thekey,thevalue}]{nstat}{data-husky-stats.dat}
\begin{table}[htbp]
  \centering
  \caption{Description of our \HuskyData{} dataset and the \VulpiData{} dataset~\cite{Vulpi2021}.
    For each class, we give the number $N$ of \SI{5}{\second} partitions, the location (Loc.), as well as the median (\,\(\tilde{\cdot}\)\,) and the \ac{IQR} of the absolute values of the linear speed \(\abs{v_x}\) and of the yaw rate \(\abs{\omega_z}\).
    Locations include San Cassiano (SC), \textit{Forêt Montmorency} (FM), and the main campus of Université Laval (UL).
  }~\label{tab:dataset}
  \begin{tabu}{lX[m,l]X[m,l]ll}
    \toprule
    Terrain & $N$ & Loc. & \(\abs{\tilde{v}_x} \left(IQR\right)\) &  \(\abs{\tilde{\omega}_z} \left(IQR\right)\) \\
    \midrule
    \multicolumn{5}{c}{\VulpiData{}~\cite{Vulpi2021}} \\
    \midrule
    \textsc{Concrete}   & \data{vterr}{CONCRETE} & \ac{SC} & \numdata{vstat}{mv-CON} (\numdata{vstat}{iv-CON}) & \numdata{vstat}{mw-CON} (\numdata{vstat}{iw-CON}) \\
    \textsc{Dirt Road}  & \data{vterr}{DIRT_ROAD} & \ac{SC} & \numdata{vstat}{mv-DIR} (\numdata{vstat}{iv-DIR}) & \numdata{vstat}{mw-DIR} (\numdata{vstat}{iw-DIR}) \\
    \textsc{Ploughed}   & \data{vterr}{PLOUGHED} & \ac{SC} & \numdata{vstat}{mv-PLO} (\numdata{vstat}{iv-PLO}) & \numdata{vstat}{mw-PLO} (\numdata{vstat}{iw-PLO}) \\
    \textsc{Unploughed} & \data{vterr}{UNPLOUGHED} & \ac{SC} & \numdata{vstat}{mv-UNP} (\numdata{vstat}{iv-UNP}) & \numdata{vstat}{mw-UNP} (\numdata{vstat}{iw-UNP}) \\
    \midrule
    \multicolumn{5}{c}{\HuskyData{} (ours)} \\
    \midrule
    \textsc{Asphalt}     & \data{nterr}{ASPHALT} & UL & \numdata{nstat}{mv-ASP} (\numdata{nstat}{iv-ASP}) & \numdata{nstat}{mw-ASP} (\numdata{nstat}{iw-ASP}) \\
    \textsc{Flooring}    & \data{nterr}{FLOORING} & UL & \numdata{nstat}{mv-FLO} (\numdata{nstat}{iv-FLO}) & \numdata{nstat}{mw-FLO} (\numdata{nstat}{iw-FLO}) \\
    \textsc{Ice}         & \data{nterr}{ICE} & UL & \numdata{nstat}{mv-ICE} (\numdata{nstat}{iv-ICE}) & \numdata{nstat}{mw-ICE} (\numdata{nstat}{iw-ICE}) \\
    \textsc{Silty Loam}  & \data{nterr}{SANDY_LOAM} & FM & \numdata{nstat}{mv-SAN} (\numdata{nstat}{iv-SAN}) & \numdata{nstat}{mw-SAN} (\numdata{nstat}{iw-SAN}) \\
    \textsc{Snow}        & \data{nterr}{SNOW} & FM & \numdata{nstat}{mv-SNO} (\numdata{nstat}{iv-SNO}) & \numdata{nstat}{mw-SNO} (\numdata{nstat}{iw-SNO}) \\
    \bottomrule
  \end{tabu}
  \vspace{-0.75\baselineskip}
\end{table}

\subsection{Implementation details}\label{sec:implementation}

As our preprocessing pipeline is inspired by \citet{Vulpi2021}, we ported their publicly available MATLAB implementation\footnote{\url{https://github.com/Ph0bi0/T_DEEP}} to Python.
To ensure a fair comparison with the baseline from \citet{Vulpi2021}, we kept the same pipeline parameters, using five folds for the cross-validation, with partitions and sample durations set at \SI{5}{\second} and \SI{1.7}{\second}, respectively.
Our pipeline implementation was then validated by running it on the \VulpiData{} dataset.
In line with our Python-based pipeline, we implemented our models with PyTorch Lightning,\footnote{\url{https://github.com/Lightning-AI/lightning}} thereby facilitating replicability and adhering to prevailing standards in deep learning.

Our \ac{CNN} and Mamba classifiers were trained, validated, and tested on the \VulpiData{} and the \HuskyData{} datasets, as well as on the combination of both datasets.
In each scenario, channel-wise normalization was performed using the minimal and maximal values derived from the training data.

For the combined dataset, downsampling each similar sensor to the smallest available frequency was necessary for the \ac{CNN} to ensure that the dimensions of the spectrograms were compatible.
Hence, we resampled our \ac{IMU} data to \SI{50}{\hertz} and the wheel data from \VulpiData{} to \SI{6.5}{\hertz}.
It is important to note that this downsampling step was not needed for Mamba, as it can handle varying frequencies without requiring adjustments.
We kept the original labels from both datasets, resulting in a classification on nine terrain types.

For our \ac{CNN}, we first applied a convolution layer with a kernel of size of one, effectively applying a \ac{MLP} individually to each frequency-time element across all channels.
Subsequently, the network sequentially applied a \ac{BN} layer%
, followed by a convolution layer with a kernel of size of three, and another \ac{BN}.
Finally, predictions were generated by applying a fully connected layer on the flattened feature maps.

For our Mamba classifier, we used two branches to treat the \ac{IMU} and wheel velocity data separately.
Each branch consists of a fully connected layer used to project the input data to a high-dimensional feature space, followed by a Mamba block.
This larger feature space enhances training stability, as the latent representation can be encoded on additional channels.
Using two branches allows the models to handle both data types independently, without requiring preprocessing steps such as padding or downsampling for input compatibility.
While multiple Mamba blocks can be stacked one after the other, we found that it did not improve the model's performance; we thus only used one Mamba block per branch.
As Mamba is a causal model, we follow \citet{Gu2023Mamba} and only keep the final hidden state of each block.
To predict the terrain type, we concatenate the final hidden states of both branches and feed them to a fully connected layer.

Our classifiers were trained on an NVIDIA RTX A6000 GPU, an AMD Ryzen Threadripper 3970X 32-core CPU, and 128~GB of RAM.
For the training phase, we utilized a further subdivision of \SI{10}{\percent} dedicated to validation, allowing us to monitor the models' performances during training.
Our source code and the \HuskyData{} dataset are publicly available in our \texttt{BorealTC} repository.\footnote{\url{https://github.com/norlab-ulaval/BorealTC}}
In addition, all training details, including hyperparameters and model checkpoints, are given in the same repository.

\section{Results}\label{sec:results}

This section presents the performance of our models on both evaluated datasets.
For all models, we reported the following metrics over all folds: the precision, the recall, the F1 score, and the accuracy.
We then analyze the influence of the train dataset size on the test error of our models.
Finally, we discuss the labeling of both datasets by comparing the latent space of both datasets.

\subsection{Models performance}\label{sec:models-perf}

To quantitatively assess the performance of our classifiers, we tested them on the \VulpiData{} and the \HuskyData{} datasets, as well as on the combination of both datasets.
All reported metrics are averaged over 5-fold cross-validation.
\cref{tab:average-metrics-vulpi} gives the performance of our \ac{CNN} and Mamba classifiers on \VulpiData{}.
Our \ac{CNN} is \SI{2.62}{\pp} more accurate than the implementation of \citet{Vulpi2021}, reported at \SI{91.5}{\percent}.
An ablation study determined that the use of a Hamming window increased the accuracy by \SI{0.6}{\pp}, while the rest of the improvement is due to better hyperparameter optimization.
It can be seen from the metrics that the \ac{CNN} outperforms Mamba on \VulpiData{}.
Since the dataset of \citet{Vulpi2021} is small, we conjecture that the \ac{CNN} has an innate advantage due to its stronger inductive bias and the direct utilization of spectrograms.

\begin{filecontents*}{metrics-baseline-CNN-1700.dat}
acc = 92.21
p-CON = 97.68
r-CON = 95.14
f-CON = 96.39
p-DIR = 90.67
r-DIR = 87.50
f-DIR = 89.06
p-PLO = 97.96
r-PLO = 97.59
f-PLO = 97.77
p-UNP = 83.22
r-UNP = 88.57
f-UNP = 85.81
\end{filecontents*}
\DTLloaddb[noheader, keys={thekey,thevalue}]{bcnn}{metrics-baseline-CNN-1700.dat}
\begin{filecontents*}{metrics-vulpi-CNN-1700-hamming.dat}
acc = 94.12
ap = 24.78
p-CON = 99.21
r-CON = 95.27
f-CON = 97.20
p-DIR = 92.40
r-DIR = 92.05
f-DIR = 92.22
p-PLO = 96.94
r-PLO = 98.96
f-PLO = 97.94
p-UNP = 88.20
r-UNP = 90.48
f-UNP = 89.32
\end{filecontents*}
\DTLloaddb[noheader, keys={thekey,thevalue}]{vcnn}{metrics-vulpi-CNN-1700-hamming.dat}
\begin{filecontents*}{metrics-vulpi-mamba-1700-optim2.dat}
acc = 86.76
ap = 23.30
p-CON = 87.13
r-CON = 83.33
f-CON = 85.19
p-DIR = 91.34
r-DIR = 83.90
f-DIR = 87.46
p-PLO = 93.93
r-PLO = 96.67
f-PLO = 95.28
p-UNP = 76.08
r-UNP = 83.93
f-UNP = 79.81
\end{filecontents*}
\DTLloaddb[noheader, keys={thekey,thevalue}]{vmam}{metrics-vulpi-mamba-1700-optim2.dat}
\begin{table}[htbp]
  \centering
  \caption{Results on the \VulpiData{} dataset.}~\label{tab:average-metrics-vulpi}
  \begin{tabu}{lX[m,l]X[m,l]X[m,l]X[m,l]}
    \toprule
    Terrain & Precision (\%) & Recall ~~(\%) & F1~score (\%) & Accuracy (\%) \\
    \midrule
    \multicolumn{5}{c}{\ac{CNN}} \\
    \midrule
    \textsc{Concrete} & \bnumdata{vcnn}{p-CON} & \bnumdata{vcnn}{r-CON} & \bnumdata{vcnn}{f-CON} &
    \multirow[c]{4}{*}{\bnumdata{vcnn}{acc}}\\
    \textsc{Dirt Road} & \bnumdata{vcnn}{p-DIR} & \bnumdata{vcnn}{r-DIR} & \bnumdata{vcnn}{f-DIR}\\
    \textsc{Ploughed} & \bnumdata{vcnn}{p-PLO} & \bnumdata{vcnn}{r-PLO} & \bnumdata{vcnn}{f-PLO}\\
    \textsc{Unploughed} & \bnumdata{vcnn}{p-UNP} & \bnumdata{vcnn}{r-UNP} & \bnumdata{vcnn}{f-UNP}\\
    \midrule
    \multicolumn{5}{c}{Mamba} \\
    \midrule
    \textsc{Concrete} & \numdata{vmam}{p-CON} & \numdata{vmam}{r-CON} & \numdata{vmam}{f-CON} &
    \multirow{4}{*}{\numdata{vmam}{acc}}\\
    \textsc{Dirt Road} & \numdata{vmam}{p-DIR} & \numdata{vmam}{r-DIR} & \numdata{vmam}{f-DIR}\\
    \textsc{Ploughed} & \numdata{vmam}{p-PLO} & \numdata{vmam}{r-PLO} & \numdata{vmam}{f-PLO}\\
    \textsc{Unploughed} & \numdata{vmam}{p-UNP} & \numdata{vmam}{r-UNP} & \numdata{vmam}{f-UNP}\\
    \bottomrule
  \end{tabu}
\end{table}

\cref{tab:average-metrics-husky} compares the performance of both classifiers on our larger dataset.
Although the \ac{CNN} demonstrates higher accuracy, the difference with Mamba is not as pronounced as in \cref{tab:average-metrics-vulpi}.
We surmise that Mamba's performance catches up to the \ac{CNN} due to the significantly larger size of \HuskyData{}, approximately nine times that of \VulpiData{}.
On the other hand, the \ac{CNN} achieves comparable performance on both datasets, albeit being less accurate on \HuskyData{}.
We attribute this lower accuracy to the higher complexity of our dataset, in terms of terrain types and input commands. %

\begin{filecontents*}{metrics-husky-CNN-1700-hamming.dat}
acc = 93.96
ap = 18.61
p-ASP = 92.98
r-ASP = 83.89
f-ASP = 88.20
p-FLO = 97.29
r-FLO = 98.70
f-FLO = 97.99
p-ICE = 97.25
r-ICE = 98.11
f-ICE = 97.68
p-SAN = 96.00
r-SAN = 97.24
f-SAN = 96.61
p-SNO = 86.84
r-SNO = 92.31
f-SNO = 89.49
\end{filecontents*}
\DTLloaddb[noheader, keys={thekey,thevalue}]{hcnn}{metrics-husky-CNN-1700-hamming.dat}
\begin{filecontents*}{metrics-husky-mamba-1700-optim2.dat}
acc = 93.68
ap = 18.35
p-ASP = 91.90
r-ASP = 85.50
f-ASP = 88.59
p-FLO = 95.46
r-FLO = 98.17
f-FLO = 96.79
p-ICE = 97.12
r-ICE = 97.36
f-ICE = 97.24
p-SAN = 95.39
r-SAN = 96.20
f-SAN = 95.79
p-SNO = 88.68
r-SNO = 91.57
f-SNO = 90.10
\end{filecontents*}
\DTLloaddb[noheader, keys={thekey,thevalue}]{hmam}{metrics-husky-mamba-1700-optim2.dat}
\begin{table}[htbp]
  \centering
  \vspace{0.5\baselineskip}
  \caption{Results on the \HuskyData{} dataset.}~\label{tab:average-metrics-husky}
  \begin{tabu}{lX[m,l]X[m,l]X[m,l]X[m,l]}
    \toprule
    Terrain & Precision (\%) & Recall ~~(\%) & F1~score (\%) & Accuracy (\%) \\
    \midrule
    \multicolumn{5}{c}{\ac{CNN}} \\
    \midrule
    \textsc{Asphalt} & \bnumdata{hcnn}{p-ASP} & \numdata{hcnn}{r-ASP} & \numdata{hcnn}{f-ASP} &
    \multirow{5}{*}{\bnumdata{hcnn}{acc}}\\
    \textsc{Flooring} & \bnumdata{hcnn}{p-FLO} & \bnumdata{hcnn}{r-FLO} & \bnumdata{hcnn}{f-FLO}\\
    \textsc{Ice} & \bnumdata{hcnn}{p-ICE} & \bnumdata{hcnn}{r-ICE} & \bnumdata{hcnn}{f-ICE}\\
    \textsc{Silty Loam} & \bnumdata{hcnn}{p-SAN} & \bnumdata{hcnn}{r-SAN} & \bnumdata{hcnn}{f-SAN}\\
    \textsc{Snow} & \numdata{hcnn}{p-SNO} & \bnumdata{hcnn}{r-SNO} & \numdata{hcnn}{f-SNO}\\
    \midrule
    \multicolumn{5}{c}{Mamba} \\
    \midrule
    \textsc{Asphalt} & \numdata{hmam}{p-ASP} & \bnumdata{hmam}{r-ASP} & \bnumdata{hmam}{f-ASP} &
    \multirow{5}{*}{\numdata{hmam}{acc}}\\
    \textsc{Flooring} & \numdata{hmam}{p-FLO} & \numdata{hmam}{r-FLO} & \numdata{hmam}{f-FLO}\\
    \textsc{Ice} & \numdata{hmam}{p-ICE} & \numdata{hmam}{r-ICE} & \numdata{hmam}{f-ICE}\\
    \textsc{Silty Loam} & \numdata{hmam}{p-SAN} & \numdata{hmam}{r-SAN} & \numdata{hmam}{f-SAN}\\
    \textsc{Snow} & \bnumdata{hmam}{p-SNO} & \numdata{hmam}{r-SNO} & \bnumdata{hmam}{f-SNO}\\
    \bottomrule
  \end{tabu}
\end{table}

\subsection{Impact of train dataset size on model performance}\label{sec:ablation}

\begin{figure}[htbp]
    \centering
    \vspace{-0.5\baselineskip}
    \includegraphics[width=1\linewidth]{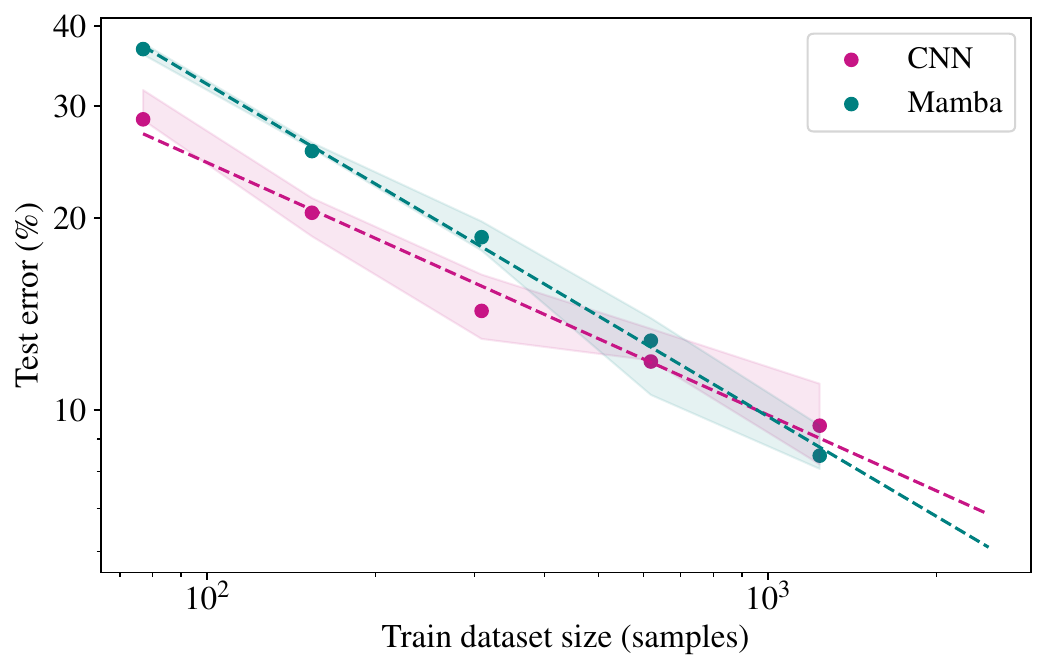}
    \caption{%
    Influence of train dataset size on the test error in log-log scale.
    Performance was assessed by combining \VulpiData{}--\HuskyData{}.
    Bands show the \ac{IQR} over 5 folds.
    }
    \label{fig:ablation}
\end{figure}

In light of Mamba's close brush with \ac{CNN} in \cref{sec:models-perf}, we performed an ablation study to examine the impact of train dataset size on test error.
To increase the amount of available data for the ablation, we used the combined dataset detailed. %
We generated several decimated training sets, with ratios of $\frac{1}{2}$, $\frac{1}{4}$, $\frac{1}{8}$ and $\frac{1}{16}$ with respect to the complete dataset.
For each subset, we applied a stratified fold strategy to maintain the same class distribution.
The test error percentage was then obtained over a 5-fold cross-validation.
\cref{fig:ablation} illustrates the impact of the train dataset size on the test error for both classifiers.
While \ac{CNN} outperforms Mamba on smaller datasets,
Mamba seems to be more accurate and could possibly surpass \ac{CNN} on larger datasets,
which aligns with our observation in \cref{sec:models-perf}.
Mamba's trend line closely follows a linear trend in log-log space, whereas the \ac{CNN} trend line is potentially sublinear.
For both models, the trend follows a typical power-law curve.
Yet, further studies may be needed to determine whether these trends hold true for larger datasets.
Overall, the results suggest that performance is predominantly limited by the amount of training data, and not by the quantity of sensor information or the learning capacity of a classifier.
Finally, we observed a worse performance compared to the separate datasets, especially for \ac{CNN}.
This could be due to overlaps between the classes in both datasets.
We investigate this behavior in the following section.

\subsection{Latent space coherence between both datasets}\label{sec:tsne}
\begin{figure}[htbp]
    \vspace{0.5\baselineskip}
    \centering
    \includegraphics[width=1\linewidth]{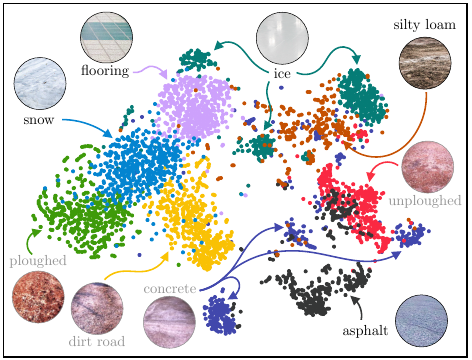}
    \caption{An illustration of class proximity using \ac{t-SNE} analysis from our \ac{CNN} classifier trained on both datasets.
    Each colored dot represents an embedding for a given class.
    Each inset illustrates a terrain class in a dataset.
    The terrains of the \HuskyData{} dataset are indicated with black labels and photos from our experiments, while the terrains of the \VulpiData{} dataset are indicated with gray labels and insets from the figures of \citet{Vulpi2021}.
    }
    \label{fig:tsne}
    \vspace{-0.75\baselineskip}
\end{figure}

When combining two \ac{TC} datasets with distinct terrain classes, it is important to ensure that each terrain is well delineated, since the data from two terrain classes may share common features, albeit being from two different datasets.
To assess this issue, we took advantage of having two different datasets, both acquired by two similar vehicles, to determine whether the labels in both datasets were properly delimited.
We used the \ac{t-SNE} technique~\cite{vanderMaaten2008} to visualize the latent space created by our models, as seen in \cref{fig:tsne}.
Specifically, \ac{t-SNE} was applied to project the features, extracted right before the fully connected layer of our \ac{CNN}, into 2-D space.
This projection allows us to visualize the proximity between labels of all classes.
The result of the \ac{t-SNE} is interpreted with our field observations and terrain descriptions from previous works on the \VulpiData{} dataset~\cite{Reina2017}.
In most cases, classes are grouped in separate clusters, which means they are well-defined.
Meanwhile, \textsc{concrete} and \textsc{ice} are each spread in three different clusters, whereas \textsc{silty loam} is dispersed between \textsc{concrete}, \textsc{ice}, and \textsc{flooring}.
This dispersion can be explained by different weather conditions or commanded body velocities.
The spread of the embeddings affects the performance when both datasets are merged, as described in the last section.
Next, we can see that the data for \textsc{asphalt}, \textsc{concrete}, and \textsc{unploughed} are in the same region, meaning that their labels coincide, as they are all hard rigid terrains.
Similarly, the embeddings for \textsc{dirt road}, \textsc{ploughed}, and \textsc{snow} are in the same zone, clustering as soft grounds.
Moreover, we noted an adjacency between \textsc{ice} and \textsc{flooring} data, two hard grounds.
This result has previously been observed by \citet{Giguere2009} with a legged robot.
Furthermore, the \textsc{ice} embeddings are close to \textsc{silty loam}, which were both slippery during our experiments.
Finally, even if the terrains are accordingly grouped with their properties, both datasets are still separated in the \ac{t-SNE} visualization.
Indeed, apart from \textsc{asphalt}, all the classes from the \VulpiData{} dataset (with lighter labels) are in the lower region of the \ac{t-SNE}, while the classes from our dataset (with darker labels) are in the upper region.
We believe that this separation indicates that both datasets are distinguishable, as they were recorded with different vehicles and different experimental procedures.
As such, our classifiers have a harder time consolidating both datasets' features, which agrees with the performance hit noted in \cref{sec:ablation}.
This observation could be verified by applying a consistent recording procedure to record various datasets on a standardized fleet of vehicles.

\section{Conclusion}\label{sec:conclusion}

In this paper, we introduced our {\HuskyData} dataset for proprioceptive-based \acf{TC}, which is one order of magnitude larger than its alternative.
Our publicly available dataset contains \ac{IMU}, motor current, and wheel odometry signals recorded with a \textit{Husky} A200 over five types of terrains, with a particular focus on boreal forests.
In particular, \HuskyData{} contains annotated data on three wintry terrain types, \textsc{snow}, \textsc{ice}, and \textsc{silty loam}, all of which are omnipresent in boreal forests.
We confirmed the capacity of a \ac{CNN} and a Mamba classifier to classify terrains on the \VulpiData{} dataset~\cite{Vulpi2021} and our dataset.
Moreover, we showed that a spectrogram-based \ac{CNN} excels on smaller \ac{TC} datasets, while Mamba performs well on increasingly larger datasets.
Additionally, a \ac{t-SNE} applied on a combined \ac{TC} dataset showed how embeddings of a type of terrain cluster with embeddings of terrains with similar properties.
However, we determined that merging two datasets does not yield a homogeneous mix of the terrain labels.
Such division could be caused by differing vehicles, sensors, and methodologies, meaning that the specificities of each dataset could have guided the classification.
Future research should aim at standardizing data acquisition procedures for \ac{TC}.
We believe gathering various datasets by applying the same experimental procedure on similar vehicles enables proper datasets for \ac{TC} without providing dataset-specific hints to classifiers.
We surmise that this need for standardized \ac{TC} experiments is aligned with the requirement for standardized terrain-aware vehicle characterization~\cite{Baril2024}, as well as models that can classify terrains in any given biome.
Finally, while we suggest that \textit{proprioception is all you need} for \ac{TC}, further research is needed to compare the performance of proprioceptive-based \ac{TC} in boreal contexts with other architectures and modalities.

\section*{Acknowledgments}

This research was supported by the Natural Sciences and Engineering Research Council of Canada (NSERC) through the grant CRDPJ 527642-18 SNOW (Self-driving Navigation Optimized for Winter).

\IEEEtriggeratref{18}
\IEEEtriggercmd{\enlargethispage{-0.1in}}

\printbibliography

\end{document}